%% file: paper.tex
\documentclass{article}
\usepackage[preprint]{spconf}
\usepackage{adjustbox, amsmath, amssymb, graphicx, multirow, booktabs}
\usepackage{algorithm}
\usepackage{algpseudocode}
\usepackage{url}
\usepackage{subfigure}
\usepackage{IEEEtrantools}
\makeatletter
\providecommand\add@text{}
\newcommand\tagaddtext[1]{%
  \gdef\add@text{#1\gdef\add@text{}}}%
\renewcommand\tagform@[1]{%
  \maketag@@@{\llap{\add@text\quad}(\ignorespaces#1\unskip\@@italiccorr)}%
}
\makeatother

\usepackage{pgfplots}
\pgfplotsset{grid style={dashed,gray}}
\pgfplotsset{compat=1.12}
\usepackage{tikz}
\usetikzlibrary{shapes.geometric}
\usetikzlibrary{shapes.misc,shadows}
\usetikzlibrary{positioning} % ...positioning nodes
\usetikzlibrary{arrows} % ...customizing arrows
\usetikzlibrary{arrows.meta}
\tikzset{%
    >={Latex[width=1mm,length=1mm]},
    base/.style = {
        rectangle, rounded corners, draw=black,
        minimum width=2cm, minimum height=.4cm,
        text centered, font=\tiny},
    acoustic_model/.style = {base, fill=red!15},
    language_model/.style = {base, fill=cyan!20},
    joint/.style = {base, fill=yellow!15},
    io/.style = {base, fill=none, draw=none, minimum width=0cm},
    data/.style = {
        rectangle, draw, align=center, left color=blue!20, right color=white,
        minimum width=0.5cm, minimum height=0.5cm},
    block/.style ={
        rectangle, thick, draw=black, align=center, fill=orange!15,
        minimum height=.3cm, minimum width=2cm, text width=2cm},
    connector/.style={-latex, font=\tiny},
    rectangle connector/.style={
        connector,
        to path={(\tikztostart) -- ++(#1,0pt) \tikztonodes |- (\tikztotarget) },
        pos=0.5
    },
}

\usepackage{color}

\copyrightnotice{\copyright\ IEEE 2021} %

\title{Exploring Heterogeneous Characteristics of Layers In ASR Models For More Efficient Training}

\name{Lillian Zhou$^*$ \qquad Dhruv Guliani$^*$ \qquad Andreas Kabel \qquad Giovanni Motta  \qquad Fran\c{c}oise Beaufays}
\address{Google LLC, Mountain View, CA, U.S.A. \\
\{lqz, dguliani, aka\}@google.com}

\begin{document}
\bstctlcite{IEEEexample:BSTcontrol}
% \ninept
%
\maketitle
\def\thefootnote{*}\footnotetext{These authors contributed equally to this work.}
\begin{abstract}

Transformer-based architectures have been the subject of research aimed at understanding their overparameterization and the non-uniform \emph{importance} of their layers. Applying these approaches to Automatic Speech Recognition, we demonstrate that the state-of-the-art Conformer models generally have multiple \emph{ambient} layers. We study the stability of these layers across runs and model sizes, propose that group normalization may be used without disrupting their formation, and examine their correlation with model weight updates in each layer. Finally, we apply these findings to Federated Learning in order to improve the training procedure, by targeting Federated Dropout to layers by importance. This allows us to reduce the model size optimized by clients without quality degradation, and shows potential for future exploration.

\end{abstract}
\input{introduction}
\input{methodology}
\input{experiments}
\input{applications}
\input{conclusions}
\input{acknowledgements}
\vfill\pagebreak
\bibliographystyle{IEEEtran}
\bibliography{refs}
\end{document}

%% file: introduction.tex
\section{Introduction}
\label{sec:intro}

Neural models for Automatic Speech Recognition (ASR)~\cite{sequence_models_in_asr} have improved to surpass conventional systems~\cite{e2e_surpasses_server}, and newer architectures, such as the Conformer~\cite{gulati_2020_conformer}, have improved quality and latency for mobile applications. These models typically contain well over 100M parameters~\cite{li_2021_better_conformer, gruestein_2021_nonstreaming_asr, li_2021_large_asr} and often require hundreds of thousands of training steps to converge. This is resource intensive~\cite{patterson2021carbon} and poses significant practical challenges when training ASR models on edge devices~\cite{dguliani_fl_asr}.

Research in deep learning has shown that neural networks benefit from being overparameterized~\cite{zhu_2018_overparam, neyshabur_2018_overparam}, and their layers exhibit non-uniform importance to the predictor's performance~\cite{bengio_2019_ambient_layers}, with \emph{ambient} layers having little impact, and \emph{critical} layers causing catastrophic changes to model quality if tampered with. Other work has found that, with transformer- and convolution-based architectures, high-quality models can be learned by training only a subset of the full model's weights~\cite{2018_rosenfeld_conv, shrivastava_2021_esn_in_asr, shen_2020_reservoir_transformers, fan_2020_structured_dropout, chatterji_2019_criticality}. We are motivated by these discoveries to study and extend the concept of ambient layers to ASR, with the hypothesis that training algorithms can be improved using the ambient characteristics of layers to inform model updates. We apply these findings to an application in Federated Learning (FL)~\cite{mcmahan_2017_fl}.

In doing so, we make the following key contributions:

\begin{itemize}
    \itemsep0em
    \item Prior work~\cite{bengio_2019_ambient_layers} explores ambient properties in convolutional models for image classification. We extend this to State-of-the-Art (SOTA) ASR models~\cite{gulati_2020_conformer}, demonstrating that ambient properties exist across model sizes, architectures, and domains.
    \item \cite{bengio_2019_ambient_layers} omits batch normalization at cost of some degradation in model quality in order to better illustrate ambient properties. We show that group normalization~\cite{wu_2018_group_norm} can be used while maintaining ambient properties.
    \item We explore model weight updates of trained models, and how they might relate to ambient properties.
    \item We show one application to a practical FL domain adaptation task, by targeting Federated Dropout~\cite{konecny_fd} using ambient properties.
\end{itemize}

%% file: methodology.tex
\section{Methodology}
\label{sec:methodology}

\subsection{Data}
\label{ssec:data}
We use the Librispeech~\cite{panayotov_2015_librispeech} corpus, containing ${\sim}960$ hours of transcribed training utterances, and ${\sim}21$ hours of evaluation audio split over 4 sets, \emph{Dev}, \emph{DevOther}, \emph{Test}, and \emph{TestOther}, with the reporting metric of~\emph{Word Error Rate} (WER). In this paper, we primarily report on the \emph{TestOther} set, though we examined the WER for all evaluation sets in our experiments and found performance to be consistent across them.

We then apply our findings to a multi-domain dataset (MD) encompassing domains of search, farfield, telephony, YouTube, etc~\cite{misra21_interspeech, recognizing_long_form}. Relevant durations are shown in Table~\ref{tab:data}. A portion of this data, termed Short-Form (SF), with an average duration of $4.6$ seconds per utterance, is the subject of our federated learning domain transfer experiments. First, we perform centralized training on the MD dataset with SF withheld (MD-SF), and then follow up with training on this withheld domain (SF) in FL simulation. The resulting model is evaluated on a disjoint test set from the SF domain.

\begin{table}[ht]
    \centering
    \small
    \begin{tabular}{c c c c}
        \toprule
        \textbf{Dataset} & \textbf{Hours} \\
        \midrule
        Multi-domain (MD) & 400k \\
        \midrule
        Short-form domain (SF) & 27k \\
        Short-form held out (MD-SF) & 373k \\
        \bottomrule
    \end{tabular}
    \caption{Durations of MD datasets.}
    \label{tab:data}
\end{table}

\subsection{Models}
\label{ssec:models}

Initial experiments analyzing ambient properties are conducted with the \textit{Non-Streaming} Conformer~\cite{gulati_2020_conformer} architecture, which is a transformer augmented with convolution layers. We use the three different sizes of models (ConformerS, ConformerM, and ConformerL) from that paper.

We look for ambient properties within the encoder, which, for these models, comprises
16 or 17 equally-sized Conformer layers (Table~\ref{tab:conformer_sizes}). These models are trained using the Librispeech corpus.

\begin{table}
    \centering
    \small
    \begin{tabular}{c c c c}
        \toprule
        \textbf{Model} & \textbf{Conf Params} & \textbf{Conf Layers} & \textbf{Total Params} \\
        \midrule
        ConformerS & 8.1M & 16 $\times$ 0.5M & 10.3M \\
        ConformerM & 25.4M & 16 $\times$ 1.6M & 30.7M \\
        ConformerL & 107.5M & 17 $\times$ 6.3M & 118.6M \\
        \bottomrule
    \end{tabular}
    \caption{Architecture of the Non-streaming Conformers.}
    \label{tab:conformer_sizes}
\end{table}

\begin{figure}[ht]
\centering
\includegraphics[width=1.0\columnwidth]{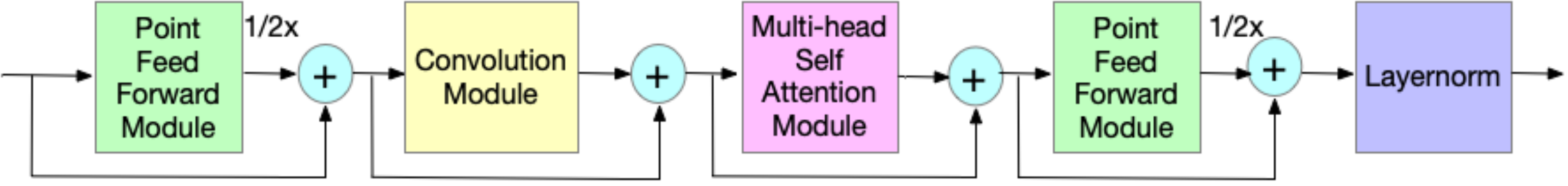}
\caption{Streaming Conformer encoder layer; from \cite{li_2021_better_conformer}.}
\label{fig:improved_conformer_layer}
\end{figure}

We expand our findings to larger-scale Multi-domain data using a similar \textit{Streaming} Conformer. This Conformer has modifications to enable latency improvements for a streaming context \cite{li_2021_better_conformer}. Specifically, the order of the convolution and multi-headed self-attention modules is reversed (Figure~\ref{fig:improved_conformer_layer}), and a frame-stacking layer is introduced as its encoder layer $4$. The additional frame-stacking increases the size of this layer relative to the rest of the encoder layers.

Of note is that in both architectures, there are residual connections around individual modules in the encoder layers, which may help to maintain predictive power when ambient layers are tampered with.

\subsection{Ambient Properties}
\label{ssec:ambient_properties}

We refer to heterogeneous characteristics~\cite{bengio_2019_ambient_layers} -- layers being ambient or critical~\cite{chatterji_2019_criticality} -- as the \emph{ambient properties} of a network. As in prior work, we identify these ambient properties through ablation studies on each model's encoder layers. As detailed above, the Conformer~\cite{gulati_2020_conformer} architectures we use contain either 16 or 17 repeated conformer layers in their encoders. Although these layers consist of multiple modules, as illustrated in Figure~\ref{fig:improved_conformer_layer}, we examine ambient characteristics at the granularity of the entire layer.

To do so, we study post-training properties of the converged model. Let $\theta$ denote a model with $D$ parametric layers $\{\theta_0,...,\theta_{D-1}\}$. Similarly, let $\theta^T$ represent the state of the model after $T$ steps of training. We wish to determine empirically which of the $D$ layers are most ambient, and which are most critical.

Specifically, we run two sets of experiments to reset each layer, and check the resulting model's accuracy (also referred to as \textit{quality}). For \emph{re-initialization} (re-init) experiments, we sequentially set one conformer layer at a time to its initial values $\theta_d^T \leftarrow \theta_d^0$. For \emph{re-randomization} (re-rand) studies, we set each layer to random values sampled from the same distribution $\theta_d^0 \thicksim P_d^0$ that the original initial values were sampled from. In both cases, an evaluation on the held-out dataset is then performed to measure WER. Layers that can be reset without significant penalty are considered ambient layers, while the remainder are critical layers.

We conduct re-initialization and re-randomization studies for different initial random seeds using SOTA architectures~\cite{gulati_2020_conformer}, reporting both ambient properties and stability across runs.

In \cite{bengio_2019_ambient_layers}, batch normalization was omitted, as it disrupts the formation of ambient properties. We run experiments using batch normalization to confirm this finding, and also try replacing it with group normalization~\cite{wu_2018_group_norm} as an alternative.

\subsection{Norms and Churn Metrics}
\label{ssec:churn_metric}

While empirically resetting layers is one way to discover ambient properties, it would be useful to find numeric properties of the trained model that would reflect which layers are ambient. Similar to \cite{bengio_2019_ambient_layers}, we check the norm of the weight matrices of our trained checkpoints across layers, but do so per module. With step $t$, layer $l$, and module $m$, we plot
$$
 \frac{\left|\mathbf W^{m,l}(t) - \mathbf W^{m,l}(0)\right|_F}{\max\limits_{l'}\left|\mathbf W^{m,l'}(t) - \mathbf W^{m,l'}(0)\right|_F}
$$ for each $m$ over $l$ for a fixed $t$, where the Frobenius norm $|\cdot|_F$ is taken over all indices other than layer and module (e.~g., attention heads).

We refer to this as the \emph{churn} metric, and compare it across Conformer layers with the ambient properties found empirically.

\subsection{Federated Domain Adaptation}
\label{ssec:federated}

Finally, to illustrate one application of ambient properties, we turn to our MD transfer learning setup. We first train a model to convergence on MD-SF, in a centralized setting. Then, we train on the unseen target domain, SF, using a trainer that simulates FL rounds across multiple clients.

To make best use of the limited amount of computational power available on edge devices, we explore using ambient properties to target memory-saving techniques during this federated portion of training. For this paper, we employ Federated Dropout~\cite{konecny_fd}, which reduces the number of parameters to train on the client, at the expense of model quality. By first empirically determining the model's ambient properties, we apply heavier dropout to the ambient layers, in order to increase the computational savings, and check the resulting WER. We compare this with an alternative setting where increased dropout is applied to critical layers.

%% file: experiments.tex
\section {Experimental Results}
\label{sec:experiments}

\subsection{Empirical Study of Ambient Properties}
\label{ssec:empirical_results}

\begin{figure}[htpb]
\centering
\includegraphics[width=.9\columnwidth]{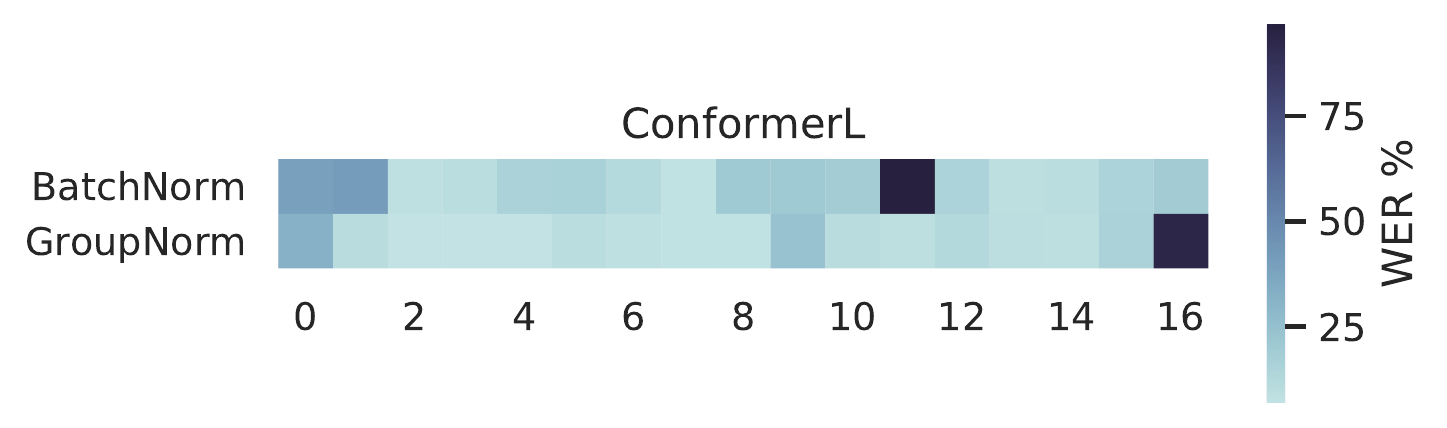}
\caption{Comparison of the same ConformerL model trained with batch normalization and group normalization, with each layer re-randomized. The model trained with group normalization was more robust to having its layers reset.}
\label{fig:bn_gn}
\end{figure}

As in \cite{bengio_2019_ambient_layers}, experiments here showed that batch normalization degrades robustness to re-randomization and re-initialization. However, we find that group normalization~\cite{wu_2018_group_norm} gives rise to comparatively stronger ambient properties, as shown in Figure~\ref{fig:bn_gn}. This suggests that group normalization can be used as an alternative to batch normalization, retaining some of the normalization benefits while still being tolerant to layer resets. This works well for FL applications, where group normalization is preferable~\cite{hsieh_2019_quagmire,reddi_2020_adaptive_fed}, and reducing training costs is especially desirable. For the rest of the paper, we use group normalization in our experiments.

\begin{figure}[htpb]
\centering
\includegraphics[width=.8\columnwidth]{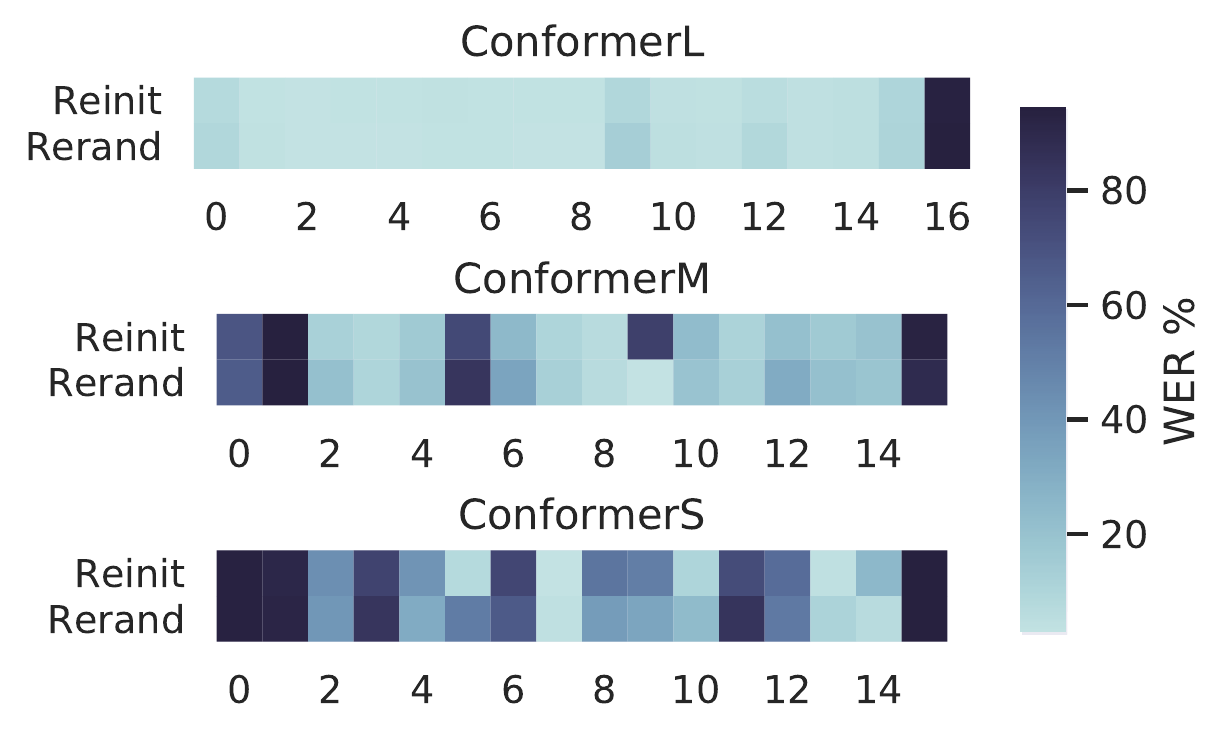}
\caption{Comparison of ambient properties following re-initialization and re-randomization across model sizes.}
\label{fig:reinit_rerand}
\end{figure}

Figure~\ref{fig:reinit_rerand} shows an example set of re-init and re-rand experiments. Typically, re-init experiments gave slightly better WER than re-rand. However, as the overall pattern of WER across layers was similar in both, we limit the remainder of our discussion to re-rand. Not only is this the more difficult task, but has deeper applications for FL, where random seeds can be transported instead of layer parameters for the most ambient layers to save on transport cost.

While ambient layers were present in all model sizes, we observed that larger models had more ambient layers, suggesting that ambient properties may be more evident in over-parameterized models.

We also performed re-rand experiments at various points throughout training. During early rounds, the ambient layers were more spread throughout the model; only later during training did the separation between ambient and critical layers become more distinct.

\begin{figure}[ht]
\centering
\includegraphics[width=.9\columnwidth]{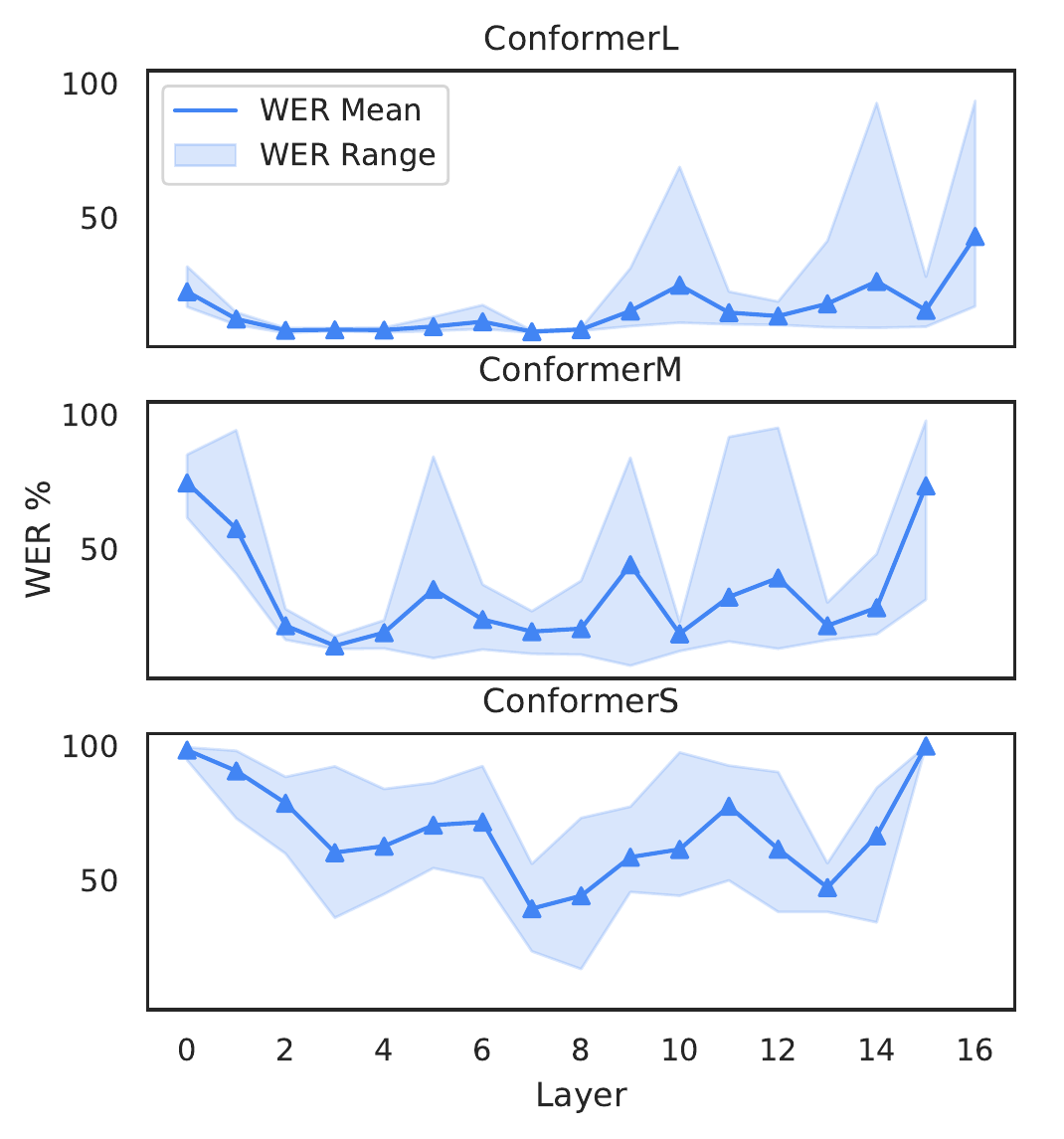}
\caption{Stability across 5 runs each of 3 model sizes. }
\label{fig:stability_comp}
\end{figure}

To check if ambient properties are consistent across runs, we trained each model five times, and show the means and range in Figure~\ref{fig:stability_comp}. We observed that larger models have a higher number of stable (illustrated by WER range), and more ambient (illustrated by mean WER) layers. For the largest model, several re-randomized layers show little deviation across runs, and give WER close to the baseline. This strengthens our hypothesis that the stability and extent of ambient properties may be related to overparameterization.

\subsection{Numerical Signatures of Ambient Properties}
\label{ssec:cost_func}

\begin{figure}[ht]
    \centering
    \subfigure[Non-streaming ConformerL trained on LibriSpeech.]{
        \includegraphics[width=0.9\columnwidth]{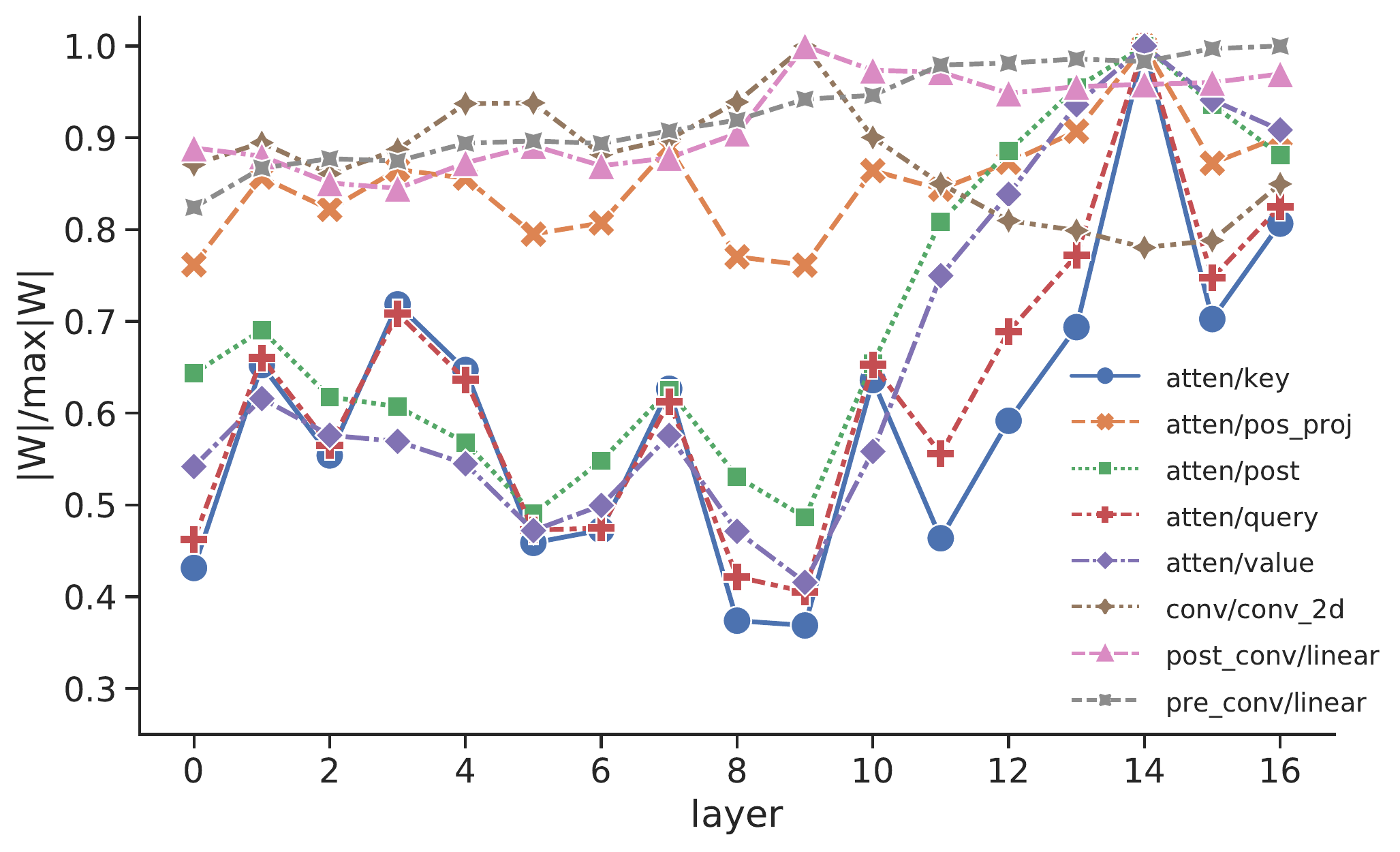}
        \label{fig:churn_conformer_lgn_70k}}
    \subfigure[Streaming Conformer trained on MD-SF.]{
        \centering
        \includegraphics[width=0.9\columnwidth]{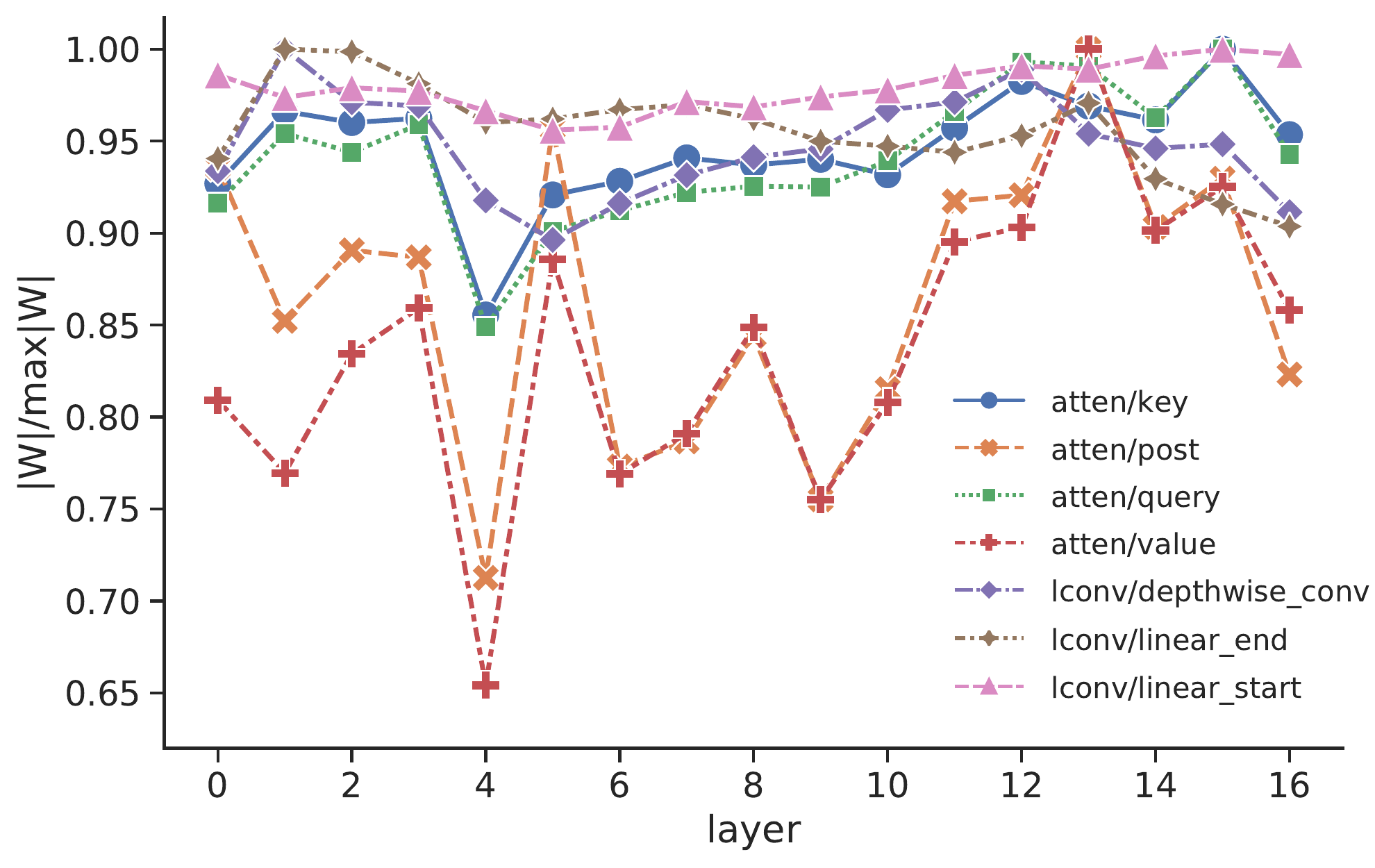}
        \label{fig:churn_canonical_conformer_450k}}
    \caption{Relative (across layers) magnitude of the change in weight matrices for different modules.}
\end{figure}

In examining the churn metric across layers in both Non-Streaming (Figure~\ref{fig:churn_conformer_lgn_70k}) and Streaming (Figure~\ref{fig:churn_canonical_conformer_450k}) Conformer models, we observed a distinct geometric pattern in attention-related modules, but not convolution-related modules. These patterns emerged during training---for earlier training phases, they were less distinct or entirely absent, in agreement with our empirical observations from ablation studies during training.

Also similarly, the emerging pattern was less distinct for ConformerM, and absent for ConformerS. Additionally, the attention modules showed greater churn in the higher layers; likewise, we observed that higher layers were more likely to be critical. This suggests that further ablation studies on a per-module basis may yield interesting results.

%% file: applications.tex
\section {Applications}
\label{sec:applications}

The following experiments were aimed at saving training costs by reducing computations on ambient layers. We explored applications in FL, where transport and edge memory are generally scarce resources~\cite{fl_field_guide}.

Federated Dropout~\cite{konecny_fd} is a technique for reducing computation and memory costs during edge training. Rather than applying flat dropout across the encoder, experiments were conducted to study whether the dropout budget could be more effectively allocated based on layer criticality.

We found that adding dropout to ambient layers gave significantly better WER results than adding dropout to critical layers. Table~\ref{tab:dropout_SF} shows the results in our domain transfer setup, where we see a $11-12\%$ relative WER improvement between applying dropout to ambient versus critical layers, given the same number of parameters dropped. Furthermore, when comparing Amb-4 $50\%$ dropout to a Flat $20\%$ dropout across the encoder, dropping from the 4 most ambient layers saves more parameters while giving the same WER. This suggests that ambient properties can find a dropout schema that improves on constant dropout across layers.

\begin{table}[ht]
    \centering
    \small
    \begin{tabular}{c c c}
        \toprule
        \textbf{Dropout} & \textbf{Params Dropped} & \textbf{WER} \\
        \midrule
        Crit-2 50\%     & 8\% & 6.9 \\
        \midrule
        Amb-2 50\%      & 9\% & 6.3 \\
        Crit-3 50\%     & 9\% & 7.0 \\
        \midrule
        Amb-3 50\%      & 10\% & 6.5 \\
        Crit-4 50\%     & 10\% & 7.3 \\
        \midrule
        Flat 20\%       & 11\% & 6.6 \\
        Amb-4 50\%      & 12\% & 6.6 \\
        \bottomrule
    \end{tabular}
    \caption{Quality of SF domain transfer task when dropout is either applied to the entire encoder, or targeted to the \emph{n} most ambient or critical layers (denoted as \emph{Amb-n} or \emph{Crit-n}).}
    \label{tab:dropout_SF}
\end{table}

We show one application of ambient properties, but many others likely exist. It may be possible to omit ambient layers entirely during training, make optimizers aware of ambient properties to improve regularization, or use ambient properties to improve neural architecture search.

%% file: conclusions.tex
\section{Conclusion}
\label{sec:conclusion}
In this work, we explored the ambient properties of ASR Conformer models and demonstrated the existence and stability of ambient layers within them. We confirmed that batch normalization is detrimental to ambient properties and illustrated the merits of group normalization instead.

Additionally, we examined weight updates on a per-module basis, showed interesting geometric signatures of attention modules in particular, and proposed a future direction of research in per-module ablation studies.

Finally, we demonstrated how ambient properties can be leveraged to improve computational efficiency during training, and proposed a potential application to Federated Learning, where such efficiency is crucial.

%% file: acknowledgements.tex
\section{Acknowledgements}
\label{sec:ack}

We would like to thank Khe Chai Sim for proof-reading and providing feedback on this paper.